\documentclass[runningheads]{llncs}
\usepackage{graphicx}
\usepackage{subfigure}
\usepackage{amsmath,amssymb} 
\usepackage{color}
\usepackage{multirow}
\usepackage[misc]{ifsym} 
\usepackage{caption}
\usepackage[colorlinks,linkcolor=blue]{hyperref}
\usepackage[width=122mm,left=12mm,paperwidth=146mm,height=193mm,top=12mm,paperheight=217mm]{geometry}
\DeclareUrlCommand{\url}{%
	\def\UrlLeft##1\UrlRight{{##1}}
}

\begin{document}

\pagestyle{headings}
\mainmatter
\newcommand*{\affmark}[1][*]{\textsuperscript{#1}}

\title{Recurrent Squeeze-and-Excitation Context Aggregation Net for Single Image Deraining} 

\titlerunning{RESCAN}

\authorrunning{Xia Li, Jianlong Wu, and et al.}

\author{Xia Li\affmark[1]\affmark[2]\affmark[3]\thanks{Equal contributions},\! Jianlong Wu\affmark[2]\affmark[3]$^{\star}$,\! Zhouchen Lin\affmark[2]\affmark[3],\! Hong Liu\affmark[1]\affmark[(\Letter)],\! and Hongbin Zha\affmark[2]\affmark[3]}


\institute{\affmark[1]Key~Laboratory~of~Machine~Perception,~Shenzhen~Graduate~School,~Peking~University\\
	\affmark[2]Key Laboratory of Machine Perception (MOE), School of EECS, Peking University\\
	\affmark[3]Cooperative Medianet Innovation Center, Shanghai Jiao Tong University \\
	\email{ \{ethanlee,jlwu1992,zlin,hongliu\}@pku.edu.cn, zha@cis.pku.edu.cn}
}

\maketitle

\begin{abstract}
	
Rain streaks can severely degrade the visibility, which causes many current computer vision algorithms fail to work.
So it is necessary to remove the rain from images.
We propose a novel deep network architecture based on deep convolutional and recurrent neural networks for single image deraining.
As contextual information is very important for rain removal, we first adopt the dilated convolutional neural network to acquire large receptive field.
To better fit the rain removal task, we also modify the network.
In heavy rain, rain streaks have various directions and shapes, which can be regarded as the accumulation of multiple rain streak layers. 
We assign different alpha-values to various rain streak layers according to the intensity and transparency by incorporating the squeeze-and-excitation block.
Since rain streak layers overlap with each other, it is not easy to remove the rain in one stage.
So we further decompose the rain removal into multiple stages.
Recurrent neural network is incorporated to preserve the useful information in previous stages and benefit the rain removal in later stages.
We conduct extensive experiments on both synthetic and real-world datasets.
Our proposed method outperforms the
state-of-the-art approaches under all evaluation metrics. Codes and supplementary material are available at our project webpage: {\url{https://xialipku.github.io/RESCAN}}.
\end{abstract}

\section{Introduction}
Rain is a very common weather in actual life. However, it can affect the visibility.
Especially in heavy rain, rain streaks from various directions accumulate and make the background scene misty,
which will seriously influence the accuracy of many computer vision systems, including video surveillance, object detection and tracking in autonomous driving, etc. 
Therefore, it is an important task to remove the rain and recover the background from rain images.
\par
Image deraining has attracted much attention in the past decade. Many methods have been proposed to solve this problem.
Existing methods can be divided into two categories, including video based approaches and single image based approaches. 
As video based methods can utilize the relationship between frames, it is relatively easy to remove rain from videos~\cite{zhang2006rain,garg2007vision,santhaseelan2015utilizing,tripathi2014removal}.
However, single image deraining is more challenging, and we focus on this task in this paper.
For single image deraining, traditional methods, such as discriminative sparse coding~\cite{luo2015removing}, low rank representation~\cite{chang2017transformed}, and the Gaussian mixture model~\cite{li2016rain}, 
have been applied to this task and they work quite well .
Recently, deep learning based deraining methods~\cite{fu2017removing,yang2017deep} receive extensive attention due to its powerful ability of feature representation.
All these related approaches achieve good performance, but there is still much space to improve.
\par
There are mainly two limitations of existing approaches.
On the one hand, according to~\cite{huang2012context,yu2015multi,chen2017fast}, spatial contextual information is very useful for deraining. However, many current methods remove rain streaks based on image patches, which neglect the contextual information in large regions.
On the other hand, as rain steaks in heavy rain have various directions and shapes, they blur the scene in different ways.
It is a common way~\cite{yang2017deep,li2017single} to 
decompose the overall rain removal problem into multiple stages,
so that we can remove rain streaks iteratively.
Since these different stages work together to remove rain streaks,
the information of deraining in previous stages is useful to guide and benefit the rain removal in later stages.
However, existing methods treat these rain streak removal stages independently and do not consider their correlations.
\par
Motivated by addressing the above two issues, we propose a novel deep network for single image deraining.
The pipeline of our proposed network is shown in Fig.~\ref{fg-rescan}.
We remove rain streaks stage by stage.
At each stage, we use the context aggregation network with multiple full-convolutional layers to remove rain streaks.
As rain streaks have various directions and shapes, each channel in our network corresponds to one kind of rain streak.
Squeeze-and-Excitation~(SE) blocks are used to assign different alpha-values to various channels according to their interdependencies in each convolution layer.
Benefited from the exponentially increasing convolution dilations, our network has a large reception field with low depth, which can help us to acquire more contextual information.
To better utilize the useful information for rain removal in previous stages,
we further incorporate the Recurrent Neural Network~(RNN) architecture with three kinds of recurrent units to guide the deraining in later stages.
We name the proposed deep network as \textbf{RE}current \textbf{S}E \textbf{C}ontext \textbf{A}ggregation \textbf{N}et~(RESCAN).
\par
Main contributions of this paper are listed as follows:

\begin{enumerate}
	\item We propose a novel unified deep network for single image deraining,
	by which we remove the rain stage by stage. Specifically, at each stage, we use the contextual dilated network to remove the rain. 
	SE blocks are used to assign different alpha-values to various rain streak layers according to their properties.
	\item To the best of our knowledge, this is the first paper to consider the correlations between different stages of rain removal. 
	By incorporating RNN architecture with three kinds of recurrent units, 
	the useful information for rain removal in previous stages can be incorporated to guide the deraining in later stages.
	Our network is suitable for recovering rain images with complex rain streaks, especially in heavy rain.
	\item Our deep network achieves superior performance compared with the state-of-the-art methods on various datasets.
\end{enumerate}

\section{Related Works}
During the past decade, many methods have been proposed to separate rain streaks and background scene from rain images.
We briefly review these related methods as follows.
\subsubsection{Video Based Methods} 
As video based methods can leverage the temporal information by analyzing the difference between adjacent frames, it is relatively easy to remove the rain from videos~\cite{barnum2010analysis,santhaseelan2015utilizing}. Garg and Nayar~\cite{garg2004detection,garg2005does,garg2007vision} propose an appearance model to describe rain streaks based on photometric properties and temporal dynamics. 
Meanwhile, Zhang et al.~\cite{zhang2006rain} exploit temporal and chromatic properties of rain in videos. 
Bossu et al.~\cite{bossu2011rain} detect the rain based on the histogram of orientation of rain streaks.
In~\cite{tripathi2014removal}, Tripathi et al. provide a review of video-based deraining methods proposed in recent years.

\subsubsection{Single Image Based Methods} 
Compared with video deraining, single image deraining is much more challenging, since there is no temporal information in images. 
For this task, traditional methods, including dictionary learning~\cite{mairal2009online}, Gaussian mixture models~(GMMs)~\cite{reynolds2000speaker}, and low-rank representation~\cite{liu2013robust}, have been widely applied.
Based on dictionary learning, 
Kang et al.~\cite{kang2012automatic} decompose high frequency parts of rain images into rain and nonrain components. 
Wang et al.~\cite{wang2017hierarchical} define a 3-layer hierarchical scheme. 
Luo et al.~\cite{luo2015removing} propose a discriminative sparse coding framework based on image patches.  
Gu et al.~\cite{gu2017joint} integrate analysis sparse representation~(ASR) and synthesis sparse representation~(SSR) to solve a variety of image decomposition problems. 
In~\cite{li2016rain}, GMM works as a prior to decompose a rain image into background and rain streaks layer. 
Chang et al.~\cite{chang2017transformed} leverage the low-rank property of rain streaks to separate two layers.
Zhu et al.~\cite{zhu2017joint} combine three different kinds of image priors. 
\par
Recently, several deep learning based deraining methods achieve promising performance. 
Fu et al.~\cite{fu2017clearing,fu2017removing} first introduce deep learning methods to the deraining problem. 
Similar to~\cite{kang2012automatic}, they also decompose rain images into low- and high-frequency parts, and then map high-frequency part to the rain streaks layer using a deep residual network. 
Yang et al.~\cite{yang2016joint,yang2017deep} design a deep recurrent dilated network to jointly detect and remove rain streaks. 
Zhang et al.~\cite{zhang2017image} use the generative adversarial network~(GAN) to prevent the degeneration of background image when  it is extracted from rain image, and utilize the perceptual loss to further ensure better visual quality. 
Li et al.~\cite{li2017single} design a novel multi-stage convolutional neural network that consists of several parallel sub-networks, each of which is made aware of different scales of rain streaks.

\section{Rain Models}
It is a commonly used rain model to 
decompose the observed rain image $\mathbf{O}$ into the linear combination of the rain-free background scene $\mathbf{B}$ and the rain streak layer $\mathbf{R}$:
\begin{equation}
	\mathbf{O}=\mathbf{B}+\mathbf{R}.
	\label{eq-linear}
\end{equation}
By removing the rain streaks layer $\mathbf{R}$ from the observed image $\mathbf{O}$, we can obtain the rain-free scene $\mathbf{B}$.
\par
Based on the rain model in Eq.~(\ref{eq-linear}), many rain removal algorithms assume that rain steaks should be sparse and have similar characters in falling directions and shapes.
However, in reality, raindrops in the air have various appearances, and occur in different distances from the camera, which leads to an irregular distribution of rain streaks. In this case, a single rain streak layer $\mathbf{R}$ is not enough to well simulate this complex situation.
\par
To reduce the complexity, we regard rain streaks with similar shape or depth as one layer. 
Then we can divide the captured rainy scene into the combination of several rain streak layers and an unpolluted background. 
Based on this, the rain model can be reformulated as follows:
\begin{equation}
	\mathbf{O}=\mathbf{B}+\sum_{i=1}^{n}\mathbf{R}^i,
	\label{eq-weight}
\end{equation}
where $\mathbf{R}^i$ represents the $i$-th rain streak layer that consists of one kind of rain streaks and $n$ is the total number of different rain streak layers.
\par
According to~\cite{kaushal2017free}, things in reality might be even worse, especially in the heavy rain situation.
Accumulation of multiple rain streaks in the air may cause attenuation and scattering, which further increases the diversity of rain streaks' brightness. 
For camera or eye visualization, the scattering causes haze or frog effects. 
This further pollutes the observed image $\mathbf{O}$. 
For camera imaging, due to the limitation of pixel number, far away rain streaks cannot occupy full pixels. 
When mixed with other things, the image will be blurry. 
To handle the issues above, we further take the global atmospheric light into consideration and assign different alpha-values to various rain streak layers according to their intensities transparencies.
We further generalize the rain model to:
\begin{equation}
	\mathbf{O}=\left(1-\sum_{i=0}^{n}\mathbf{\alpha}_i\right)\mathbf{B}+\alpha_{0}\mathbf{A}+\sum_{i=1}^{n}\mathbf{\alpha}_i \mathbf{R}^i, \ s.t. \ \alpha_{i} \ge 0, \ \sum_{i=0}^{n} \alpha_{i} \leq 1,
	\label{eq-alpha}
\end{equation}
where $\mathbf{A}$ is the global atmospheric light, $\alpha_0$ is the scene transmission, $\mathbf{\alpha}_i$~$(i=1,\cdots,n)$ indicates the brightness of a rain streak layer or a haze layer.

\section{Deraining Method}
Instead of using decomposition methods with artificial priors to solve the problem in Eq.~(\ref{eq-alpha}), we intend to learn a function $f$ that directly maps observed rain image $\mathbf{O}$ to rain streak layer $\mathbf{R}$, since $\mathbf{R}$ is sparser than $\mathbf{B}$ and has simpler texture.
Then we can subtract $\mathbf{R}$ from $\mathbf{O}$  to get the rain-free scene $\mathbf{B}$. 
The function $f$ above can be represented as a deep neural network and be learned by optimizing the loss function $\left \| f\left(\mathbf{O}\right) -\mathbf{R} \right \|_F^2$. 
\par
\begin{figure}[t]
	\centering
	\includegraphics[width=0.97\textwidth]{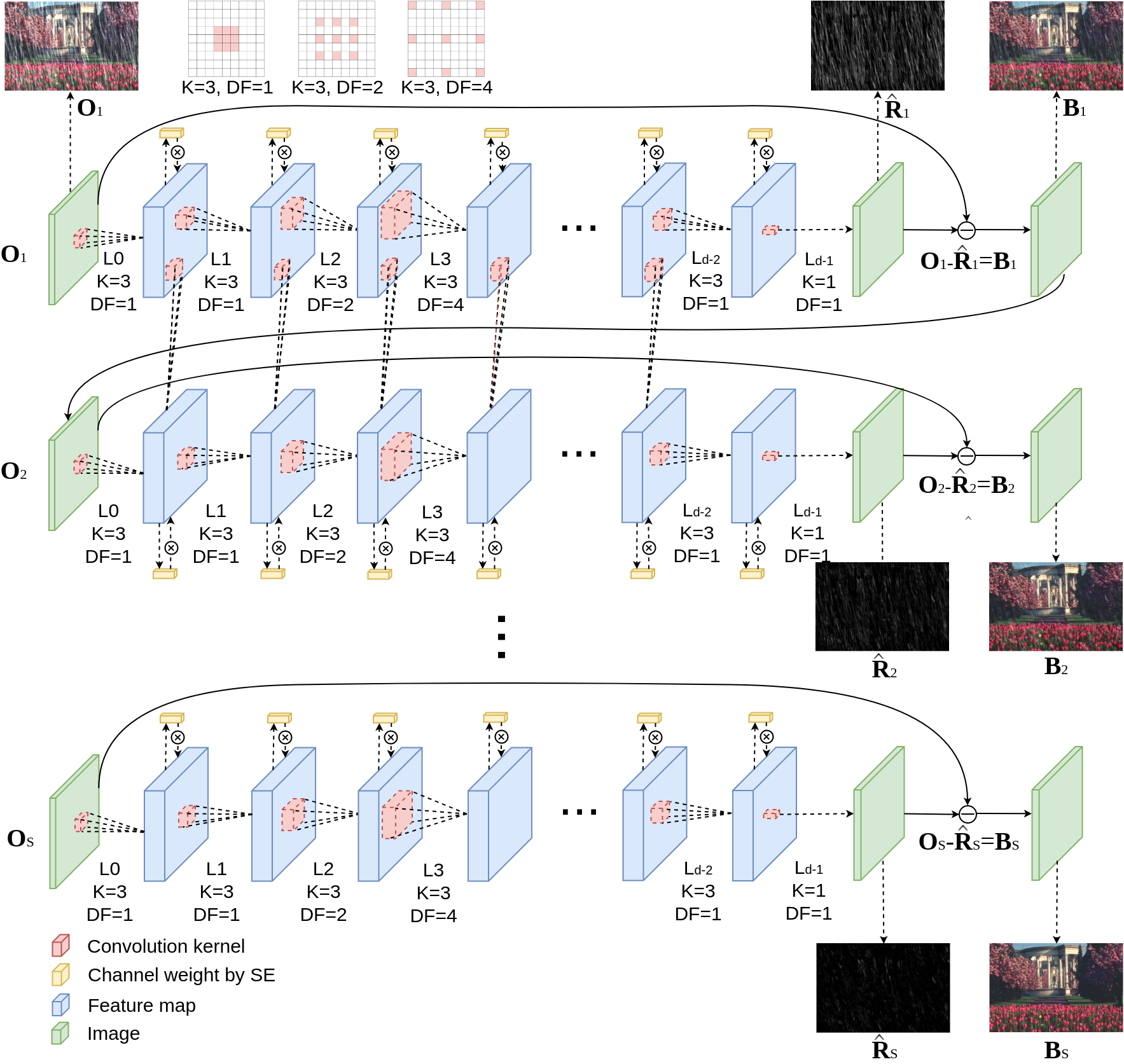}
	\caption{The unfolded architecture of RESCAN. \textit{K} is the convolution kernel size, \textit{DF} indicates the dilation factor, and \textit{S} represents the number of stages.}
	\label{fg-rescan}
\end{figure}

Based on the motivation above, we propose the \textbf{RE}current \textbf{S}E \textbf{C}ontext \textbf{A}ggregation \textbf{N}et~(RESCAN) for image deraining.
The framework of our network is presented in Fig.~\ref{fg-rescan}.
We remove rain streaks stage by stage.
At each stage, we use the context aggregation network with SE blocks to remove rain streaks.
Our network can deal with rain streaks of various directions and shapes, and each feature map in the network corresponds to one kind of rain streak.
Dilated convolution used in our network can help us have a large reception field and acquire more contextual information.
By using SE blocks, we can assign different alpha-values to various feature maps according to their interdependencies in each convolution layer.
As we remove the rain in multiple stages, useful information for rain removal in previous stages can guide the learning in later stages.
So we incorporate the RNN architecture with memory unit to make full use of the useful information in previous stages.
\par
In the following, we first describe the baseline model, and then define the recurrent structure, which lifts the model's capacity by iteratively decomposing rain streaks with different characteristics.

\begin{figure}[!t]
	\centering
	\includegraphics[width=0.85\textwidth]{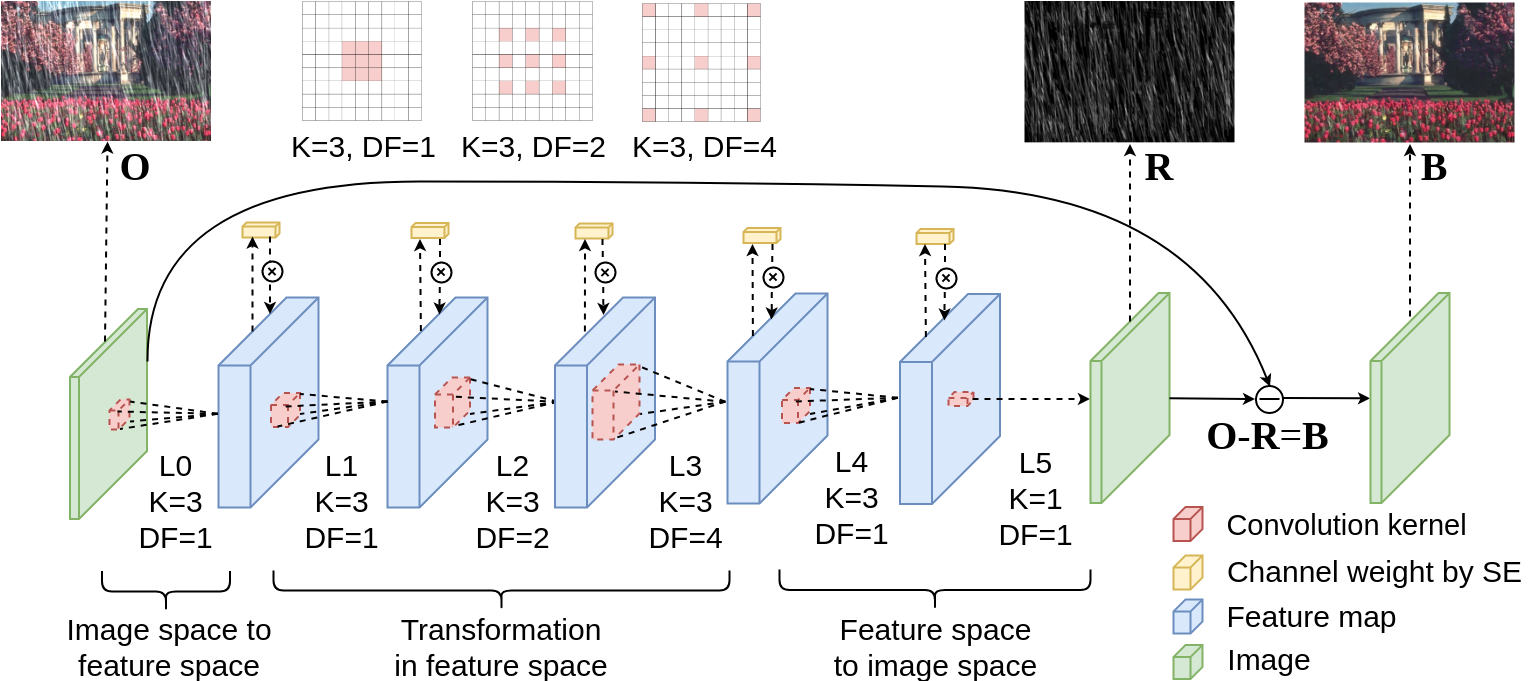}
	\caption{The architecture of \textbf{S}E \textbf{C}ontext \textbf{A}ggregation \textbf{N}etwork~(SCAN).}
	\label{fg-scan}
\end{figure}

\subsection{SE Context Aggregation Net}
The base model of RESCAN is a forward network without recurrence. We implement it by extending \textbf{C}ontext \textbf{A}ggregation \textbf{N}et~(CAN)~\cite{yu2015multi,chen2017fast} with Squeeze-and-Excitation~(SE) blocks~\cite{hu2017squeeze}, and name it as \textbf{S}E \textbf{C}ontext \textbf{A}ggregation \textbf{N}et~(SCAN). 
\par
Here we provide an illustration and a further specialization of SCAN. 
We schematically illustrate SCAN in Fig.~\ref{fg-scan}, which is a full-convolution network. 
In Fig.~\ref{fg-scan}, we set the depth $d=6$. 
Since a large receptive field is very helpful to acquire much contextual information, dilation is adopted in our network.
For layers $L_{1}$ to $L_{3}$, the dilation increases from 1 to 4 exponentially, which leads to the exponential growth in receptive field of every elements.
As we treat the first layer as an encoder to transform an image to feature maps, and the last two layers as a decoder to map reversely,
we do not apply dilation for layers $L_{0}$, $L_{4}$ and $L_{5}$. 
Moreover, we use $3\times 3$ convolution for all layers before $L_{5}$. To recover RGB channels of a color image, or gray channel for a gray scale image, we adopt the $1\times 1$ convolution for the last layer $L_{5}$. Every convolution operation except the last one is followed by a nonlinear operation. The detailed architecture of SCAN is summarized in Table 1. Generally, for a SCAN with depth $d$, the receptive field of elements in the output image equals to $\left(2^{d-2}+3\right)^{2}$.
\par
\begin{table}[t]
	\begin{center}
		\caption{The detailed architecture of SCAN. $d$ is the depth of network.}
		\label{table-net}
		\begin{tabular}{cccccccc}
			\hline\noalign{\smallskip}
			Layer & 0 & 1 & 2 & ... & $d-3$ & $d-2$ & $d-1$ \\
			\noalign{\smallskip}
			\hline
			\noalign{\smallskip}
			Convolution     & $3\times 3$& $3\times 3$& $3\times 3$& ...        & $3\times 3$& $3\times 3$ & $1\times 1$\\
			Dilation        & 1          & 1          & 2          & ...        & $2^{d-4} $ & 1           & 1          \\
			NonLinear       & Yes        & Yes        & Yes        & Yes        & Yes        & Yes         & No         \\
			SE block        & Yes        & Yes        & Yes        & Yes        & Yes        & Yes         & No         \\
			Receptive field & $3\times 3$& $5\times 5$& $9\times 9$& ...        & $\left(2^{d-2}+1\right)^2$ & $\left(2^{d-2}+3\right)^2$ & $\left(2^{d-2}+3\right)^2$ \\
			\hline
		\end{tabular}
	\end{center}
\end{table}

For feature maps, we regard each channel of them as the embedding of a rain streak layer $\mathbf{R}^i$. 
Recall in Eq.~(\ref{eq-alpha}), we assign different alpha-values $\mathbf{\alpha}_i$ to different rain steak layers $\mathbf{R}^i$. 
Instead of setting fixed alpha-values $\mathbf{\alpha}_i$ for each rain layer, we update the alpha-values for embeddings of rain streak layers in each network layer. 
Although the convolution operation implicitly introduces weights for every channel, these implicit weights are not specialized for every image. 
To explicitly import weight on each network layer for each image, we extend each basic convolution layer with Squeeze-and-Excitation~(SE) block~\cite{hu2017squeeze},
which computes normalized alpha-value for every channel of each item. 
By multiplying alpha-values learned by SE block, feature maps computed by convolution are re-weighted explicitly. 
\par
An obvious difference between SCAN and former models~\cite{fu2017removing,zhang2017image} is that SCAN has no batch normalization~(BN)~\cite{ioffe2015batch} layers. 
BN is widely used in training deep neural network, as it can reduce internal covariate shift of feature maps. By applying BN, each scalar feature is normalized and has zero mean and unit variance. These features are independent with each other and have the same distribution. 
However, in Eq.~(\ref{eq-weight}) rain streaks in different layers have different distributions in directions, colors and shapes, 
which is also true for each scalar feature of different rain streak layers.
Therefore, BN contradicts with the characteristics of our proposed rain model. 
Thus, we remove BN from our model. 
Experimental results in Section~\ref{sec-exp} show that this simple modification can substantially improve the performance. 
Furthermore, since BN layers keep a normalized copy of feature maps in GPU, removing it can greatly reduce the demand on GPU memory. 
For SCAN, we can save approximately $40\%$ of memory usage during training without BN.
Consequently, we can build a larger model with larger capacity, or increase the mini-batch size to stabilize the training process.
\subsection{Recurrent SE Context Aggregation Net}\label{sec-rescan}
As there are many different rain streak layers and they overlap with each other, it is not easy to remove all rain streaks in one stage. So we incorporate the recurrent structure to decompose the rain removal into multiple stages. 
This process can be formulated as:
\begin{align}
	\mathbf{O}_1 &= \mathbf{O}, \\
	\mathbf{{R}}_s & = f_{CNN}\left(\mathbf{O}_s\right) , 1 \leq s \leq S, \label{eq-iter} \\
	\mathbf{O}_{s+1} &= \mathbf{O}_{s} - \mathbf{{R}}_s, 1 \leq s \leq S ,\\
	\mathbf{R} & = \sum_{s=1}^{S}\mathbf{{R}}_s,
\end{align}
where $S$ is the number of stages, $\mathbf{{R}}_s$ is the output of the $s$-th stage, 
and $\mathbf{O}_{s+1}$ is the intermediate rain-free image after the $s$-th stage.
\par
The above model for deraining has been used in~\cite{yang2017deep,li2017single}.
However, recurrent structure used in their methods~\cite{yang2017deep,li2017single} can only be regarded as the simple cascade of the same network. 
They just use the output images of last stage as the input of current stage and do not 
consider feature connection among these stages.
As these different stages work together to remove the rain, 
input images of different stages $\{\mathbf{O}_1,\mathbf{O}_2,\cdots,\mathbf{O}_S \} $ can be regarded as a temporal sequence of rain images with decreasing levels of rain streaks interference.
It is more meaningful to investigate the recurrent connections between features of different stages rather than only using the recurrent structure.
So we incorporate recurrent neural network~(RNN)~\cite{mandic2001recurrent} with memory unit to better make use of information in previous stages
and guide feature learning in later stages.
\par
In our framework, Eq.~(\ref{eq-iter}) can be further reformulated as:
\begin{equation}
	\mathbf{{R}}_s = \sum_{i=1}^{n} \alpha_{i}  \mathbf{\bar{R}}^{i}_{s} = f_{CNN+RNN}\left(\mathbf{O}_s, x_{s-1}\right) , 1 \leq s \leq S,
\end{equation}
where $\mathbf{\bar{R}}^{i}_{s}$ is the decomposed $i$-th rain streak layer of the $s$-th stage and 
$x_{s-1}$ is the hidden state of the ($s-1$)-th stage.
Consistent with the rain model in Eq.~(\ref{eq-alpha}), $\mathbf{{R}}_s$ is computed by summing $\mathbf{\bar{R}}^{i}_{s}$ with different alpha-values.

\par

For the deraining task, we further explore three different recurrent unit variants, including ConvRNN, ConvGRU~\cite{cho2014learning}, and ConvLSTM~\cite{zaremba2014recurrent}. 
Due to the space limitation, we only present the details of ConvGRU~\cite{cho2014learning} in the following. For other two kinds of recurrent units, please refer to the supplement materials on our webpage.
%
\subsubsection{ConvGRU}
Gated Recurrent Units~(GRU)~\cite{cho2014learning} is a very popular recurrent unit in sequential models. Its convolutional version ConvGRU is adopted in our model.
Denote $x^{j}_{s}$ as the feature map of the $j$-th layer in the $s$-th stage, and it can be computed  based on the $x^{j}_{s-1}$~(feature map in the same layer of the previous stage) and $x^{j-1}_{s}$~(feature map in the previous layer of the same stage):
\begin{align}
	z^{j}_s &= \sigma\left( W^{j}_z\circledast x^{j-1}_s + U^{j}_z\circledast x^{j}_{s-1} + b^{j}_z \right), \\
	r^{j}_s &= \sigma\left( W^{j}_r\circledast x^{j-1}_s + U^{j}_r\circledast x^{j}_{s-1} + b^{j}_r \right), \\
	n^{j}_s &= \tanh\left( W^{j}_n\circledast x^{j-1}_s + U^{j}_n\circledast\left(r^{j}_s\odot x^{j}_{s-1}\right) + b^{j}_n \right), \\
	x^{j}_{s} &= \left(1 - z^{j}_s\right) \odot x^{j}_{s-1} + z^{j}_s \odot n^{j}_s,
	\label{eq-gru}
\end{align}
where $\sigma$ is the sigmoid function $\sigma\left(x\right)=1/\left(1+\exp\left(-x\right)\right)$ and $\odot$ denotes element-wise multiplication. 
$W$ is the dilated convolutional kernel, and $U$ is an ordinary kernel with a size $3\times 3$ or $1\times 1$.
Compared with a convolutional unit, ConvRNN, ConvGRU and ConvLSTM units will have twice, three times and four times parameters, respectively.


\subsection{Recurrent Frameworks}
We further examine two frameworks to infer the final output. Both of them use $\mathbf{O}_{s}$ and previous states as input for the $s$-th stage, but they output different images. 
In the following, we describe the additive prediction and the full prediction in detail, together with their corresponding loss functions. 

\subsubsection{Additive Prediction}
The additive prediction is widely used in image processing. 
In each stage, the network only predicts the residual between previous predictions and the ground truth. 
It incorporates previous feature maps and predictions as inputs, which can be formulated as:
\begin{align}
	\mathbf{R}_s &= f\left(\mathbf{O}_{s}, x_{s-1}\right), \\
	\mathbf{O}_{s+1} &= \mathbf{O} - \sum_{j=1}^{s}\mathbf{R}_{j} = \mathbf{O}_s - \mathbf{R}_s,
	\label{eq-add}
\end{align}
where $x_{s-1}$ represents previous states as in Eq.~(\ref{eq-gru}). For this framework, the loss functions are chosen as follows:
\begin{equation}
	L\left(\mathbf{\Theta}\right) = \sum_{s=1}^{S}\left\|\sum_{j=1}^{s}\mathbf{R}_j - \mathbf{R}\right\|^2_F,
\end{equation}
where $\mathbf{\Theta}$ represents the network's parameters.

\subsubsection{Full Prediction}
The full prediction means that in each stage, we predict the whole rain streaks $\mathbf{R}$. This approach can be formulated as:
\begin{align}
	\hat{\mathbf{R}}_s = f\left(\hat{\mathbf{O}}_s, x_{s-1}\right), \\
	\hat{\mathbf{O}}_{s+1} = \mathbf{O} - \hat{\mathbf{R}}_{s},
\end{align}
where $\hat{\mathbf{R}}_s$ represents the predicted full rain streaks in the $s$-th stage, and $\hat{\mathbf{O}}_{s+1}$ equals to $\mathbf{B}$ plus remaining rain streaks. The corresponding loss function is:
\begin{equation}
	L\left(\mathbf{\Theta}\right) = \sum_{s=1}^{S}\left\|\hat{\mathbf{R}}_s - \mathbf{R}\right\|^2_F,
\end{equation}

\section{Experiments}\label{sec-exp}
In this section, we present details of experimental settings and quality measures used to evaluate the proposed SCAN and RESCAN models. 
We compare the performance of our proposed methods with the state-of-the-art methods on both synthetic and real-world datasets.

\subsection{Experiment Settings}
\subsubsection{Synthetic Dataset}
Since it is hard to obtain a large dataset of clean/rainy image pairs from real-world data, we first use synthesized rainy datasets to train the network. 
Zhang et al.~\cite{zhang2017image} synthesize 800 rain images~(Rain800) from randomly selected outdoor images, and split
them into testing set of 100 images and training set of 700 images.
Yang et al.~\cite{yang2017deep} collect and synthesize 3 datasets, Rain12, Rain100L and Rain100H. 
We select the most difficult one, Rain100H, to test our model. 
It is synthesized with the combination of five streak directions, which makes it hard to effectively remove all rain streaks. 
There are $1,800$ synthetic image pairs in Rain100H, and $100$ pairs are selected as the testing set.

\subsubsection{Real-world Dataset}
Zhang et al.~\cite{zhang2017image} and Yang et al.~\cite{yang2017deep} also provide many real-world rain images. These images are diverse in terms of content as well as intensity and orientation of rain streaks. 
We use these datasets for objective evaluation.

\subsubsection{Training Settings}
In the training process, we randomly generate $100$ patch pairs with a size of $64\times 64 $ from every training image pairs. 
The entire network is trained on an Nvidia 1080Ti GPU based on Pytorch. 
We use a batch size of $64$ and set the depth of SCAN as $d=7$  with the receptive field size $35\times 35$. 
For the nonlinear operation, we use leaky ReLU~\cite{maas2013rectifier} with $\alpha=0.2$. 
For optimization, the ADAM algorithm~\cite{kingma2014adam} is adopted with a start learning rate $5\times10^{-3}$. 
During training, the learning rate is divided by $10$ at $15,000$ and $17,500$ iterations.

\subsubsection{Quality Measures}
To evaluate the performance on synthetic image pairs, we adopt two commonly used metrics, including peak signal to noise ratio~(PSNR)~\cite{huynh2008scope} and structure similarity index~(SSIM)~\cite{wang2004image}.
Since there are no ground truth rain-free images for real-world images, the performance on the real-world dataset can only be evaluated visually. 
We compare our proposed approach with five state-of-the-art methods, including image decomposition~(ID)~\cite{kang2012automatic}, discriminative sparse coding~(DSC)~\cite{luo2015removing}, layer priors~(LP)~\cite{li2016rain}, DetailsNet~\cite{fu2017removing}, and joint rain detection and removal~(JORDER)~\cite{yang2017deep}.

\subsection{Results on Synthetic Data}

\begin{figure}[t]
	\begin{minipage}{0.19\textwidth}
		\includegraphics[width=\textwidth]{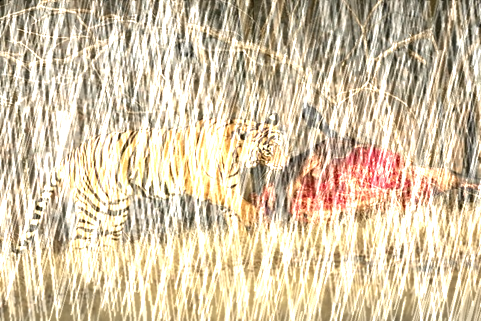}
		\caption*{$\mathbf{O}$\protect\\ 7.71/0.2334}
	\end{minipage}
	\begin{minipage}{0.19\textwidth}
		\includegraphics[width=\textwidth]{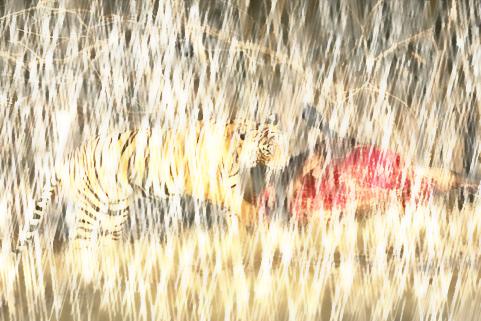}
		\caption*{\centerline{ID~\cite{kang2012automatic}}\protect\\ \centerline{7.96/0.2568}}
	\end{minipage}
	\begin{minipage}{0.19\textwidth}
		\includegraphics[width=\textwidth]{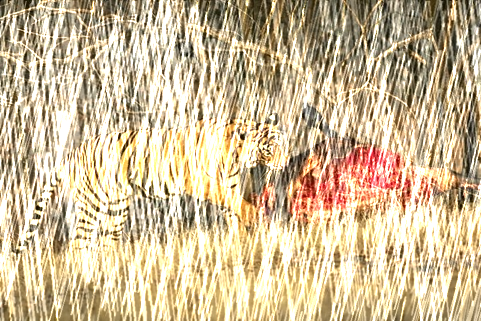}
		\caption*{\centerline{DSC~\cite{luo2015removing}}\protect\\ \centerline{8.28/0.2316}}
	\end{minipage}
	\begin{minipage}{0.19\textwidth}
		\includegraphics[width=\textwidth]{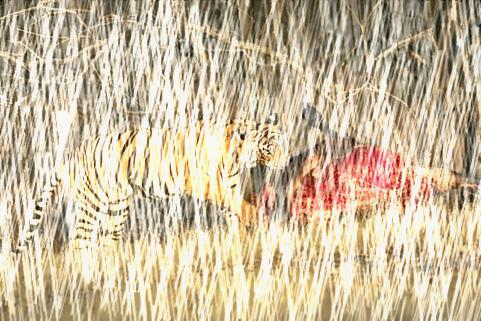}
		\caption*{\centerline{LP~\cite{li2016rain}}\protect\\ \centerline{8.60/0.2448}}
	\end{minipage}
	\begin{minipage}{0.19\textwidth}
		\includegraphics[width=\textwidth]{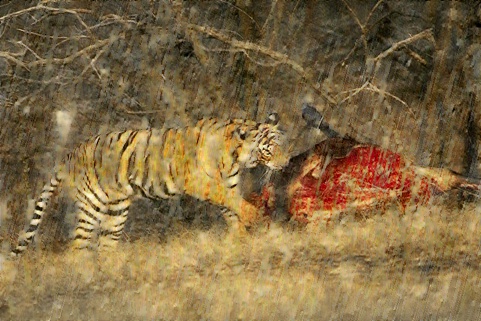}
		\caption*{\centerline{DetailsNet~\cite{fu2017removing}}\protect\\ \centerline{20.23/0.5566}}
	\end{minipage}
	
	\begin{minipage}{0.19\textwidth}
		\includegraphics[width=\textwidth]{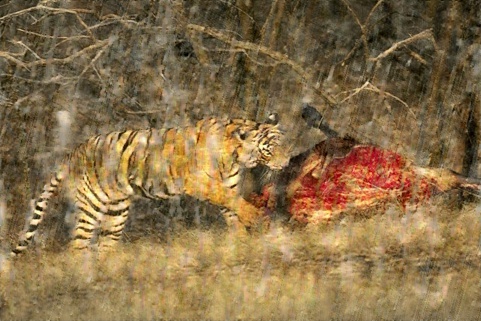}
		\caption*{\centerline{JORDER~\cite{yang2017deep}}\protect\\ \centerline{19.40/0.5535}}
	\end{minipage}
	\begin{minipage}{0.19\textwidth}
		\includegraphics[width=\textwidth]{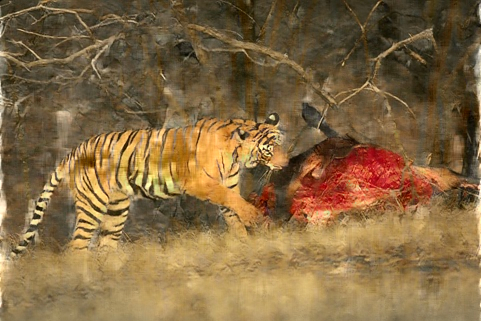}
		\caption*{\centerline{JORDER-R~\cite{yang2017deep}}\protect\\ \centerline{20.19/0.6108}}
	\end{minipage}
	\begin{minipage}{0.19\textwidth}
		\includegraphics[width=\textwidth]{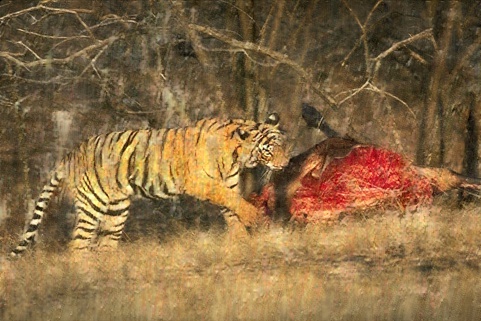}
		\caption*{\centerline{SCAN}\protect\\ \centerline{\underline{21.32}/\underline{0.6271}}}
	\end{minipage}
	\begin{minipage}{0.19\textwidth}
		\includegraphics[width=\textwidth]{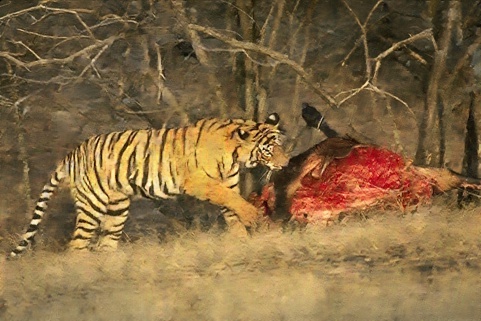}
		\caption*{\centerline{RESCAN}\protect\\ \centerline{\textbf{23.47}/\textbf{0.7035}}}
	\end{minipage}
	\begin{minipage}{0.19\textwidth}
		\includegraphics[width=\textwidth]{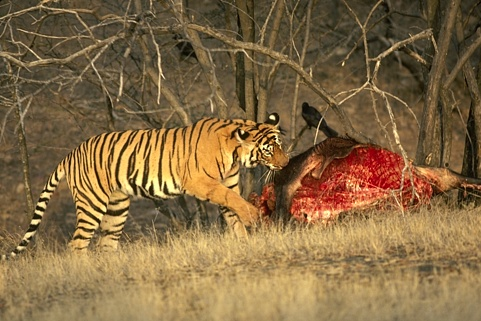}
		\caption*{$\mathbf{B}$\protect\\ Inf/1}
	\end{minipage}
	
	\begin{minipage}{0.19\textwidth}
		\includegraphics[width=\textwidth]{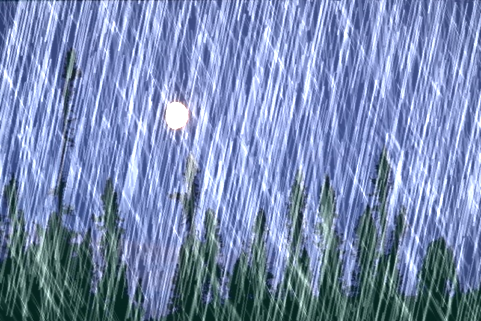}
		\caption*{$\mathbf{O}$\protect\\ 9.73/0.1149}
	\end{minipage}
	\begin{minipage}{0.19\textwidth}
		\includegraphics[width=\textwidth]{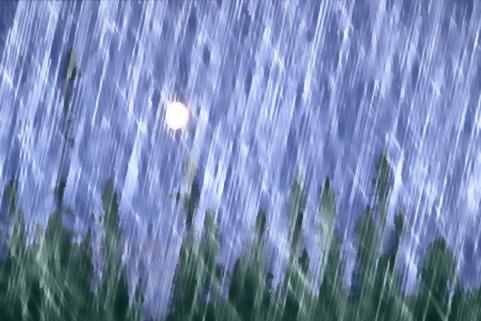}
		\caption*{\centerline{ID~\cite{kang2012automatic}}\protect\\ \centerline{10.42/0.2635}}
	\end{minipage}
	\begin{minipage}{0.19\textwidth}
		\includegraphics[width=\textwidth]{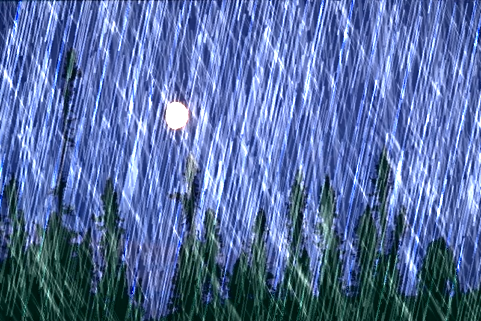}
		\caption*{\centerline{DSC~\cite{luo2015removing}}\protect\\ \centerline{11.87/0.1238}}
	\end{minipage}
	\begin{minipage}{0.19\textwidth}
		\includegraphics[width=\textwidth]{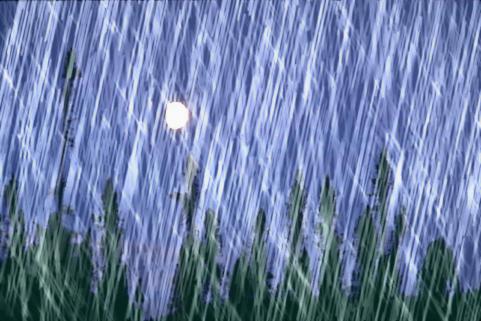}
		\caption*{\centerline{LP~\cite{li2016rain}}\protect\\ \centerline{11.26/0.1762}}
	\end{minipage}
	\begin{minipage}{0.19\textwidth}
		\includegraphics[width=\textwidth]{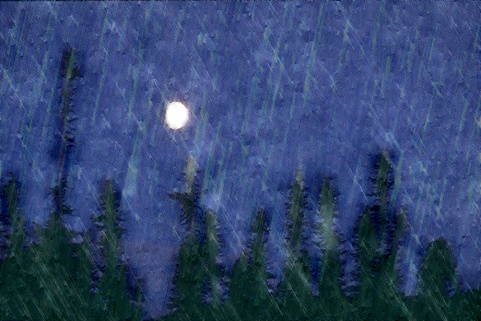}
		\caption*{\centerline{DetailsNet~\cite{fu2017removing}}\protect\\ \centerline{23.62/0.5223}}
	\end{minipage}
	
	\begin{minipage}{0.19\textwidth}
		\includegraphics[width=\textwidth]{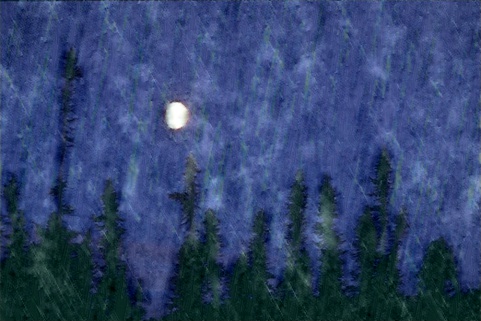}
		\caption*{\centerline{JORDER~\cite{yang2017deep}}\protect\\ \centerline{23.87/0.5489}}
	\end{minipage}
	\begin{minipage}{0.19\textwidth}
		\includegraphics[width=\textwidth]{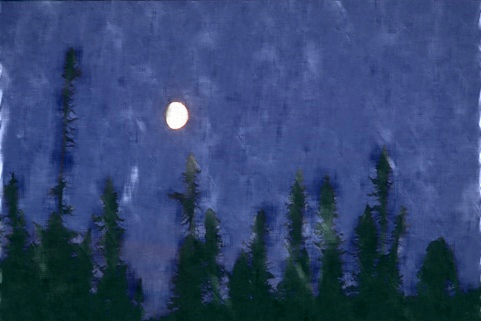}
		\caption*{\centerline{JORDER-R~\cite{yang2017deep}}\protect\\ \centerline{\underline{26.25}/\underline{0.7972}}}
	\end{minipage}
	\begin{minipage}{0.19\textwidth}
		\includegraphics[width=\textwidth]{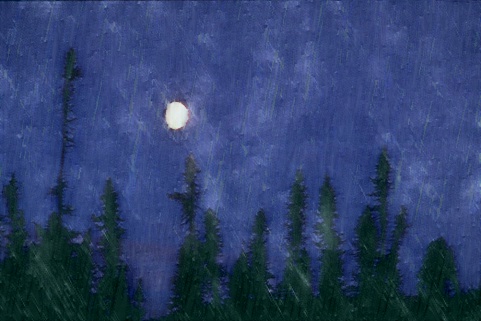}
		\caption*{\centerline{SCAN}\protect\\ \centerline{26.24/0.6831}}
	\end{minipage}
	\begin{minipage}{0.19\textwidth}
		\includegraphics[width=\textwidth]{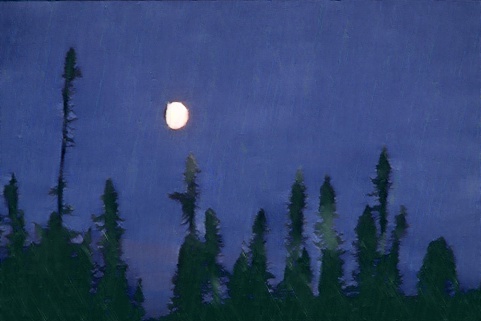}
		\caption*{\centerline{RESCAN} \protect\\ \centerline{\textbf{30.78}/\textbf{0.8807}}}
	\end{minipage}
	\begin{minipage}{0.19\textwidth}
		\includegraphics[width=\textwidth]{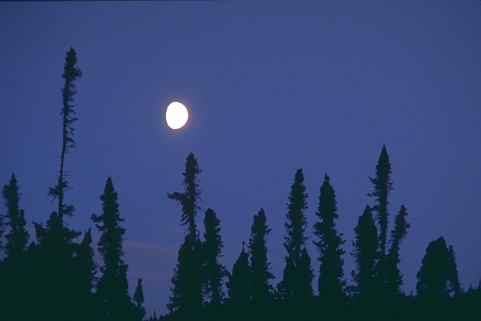}
		\caption*{$\mathbf{B}$\protect\\ Inf/1}
	\end{minipage}
	\caption{Results of various methods on synthetic images.  \textbf{Best viewed at screen!}}
	\label{fig-syn}
\end{figure}

Table~\ref{table-syn} shows results of different methods on the Rain800 and the Rain100H datasets. 
We can see that                                                                                                                               our RESCAN considerably outperforms other methods in terms of both PSNR and SSIM on these two datasets. 
It is also worth noting that our non-recurrent network SCAN can even outperform JORDER and DetailsNet, and is slightly superior to JORDER-R, the recurrent version of JORDER.
This shows the high capacity behind SCAN's shallow structure. 
Moreover, by using RNN to gradually recover the full rain streak layer $\mathbf{R}$, RESCAN further improves the performance.
\par

\begin{table}[t]
	\begin{center}
		\caption{Quantitative experiments evaluated on two synthetic datasets. 
			Best results are marked in bold and the second best results are underlined.}
		\label{table-syn}
		\begin{tabular}{cccccccc}
			\hline\noalign{\smallskip}
			Dataset &\multicolumn{2}{c}{Rain800} &\multicolumn{2}{c}{Rain100H} \\
			\noalign{\smallskip}
			\hline
			\noalign{\smallskip}
			Meassure& PSNR  & SSIM   & PSNR  & SSIM   \\
			\noalign{\smallskip}
			\hline
			\noalign{\smallskip}
			ID~\cite{kang2012automatic}     & 18.88 & 0.5832 & 14.02 & 0.5239 \\
			DSC~\cite{luo2015removing}      & 18.56 & 0.5996 & 15.66 & 0.4225 \\
			LP~\cite{li2016rain}            & 20.46 & 0.7297 & 14.26 & 0.5444 \\
			DetailsNet~\cite{fu2017removing}& 21.16 & 0.7320 & 22.26 & 0.6928\\
			JORDER~\cite{yang2017deep}      & 22.24 & 0.7763 & 22.15 & 0.6736 \\
			JORDER-R~\cite{yang2017deep}    & 22.29 & 0.7922 & 23.45 & \underline{0.7490} \\
			\noalign{\smallskip}
			\hline
			\noalign{\smallskip}
			SCAN      					    & \underline{23.45} & \underline{0.8112} & \underline{23.56} & 0.7456 \\
			RESCAN                          & \textbf{24.09} & \textbf{0.8410} & \textbf{26.45} & \textbf{0.8458} \\
			\hline
		\end{tabular}
	\end{center}
	\vspace{-2mm}
\end{table}

To visually demonstrate the improvements obtained by the proposed methods, we present results on several difficult sample images in Fig.~\ref{fig-syn}. 
Please note that we select difficult sample images to show that our method can outperform others especially in difficult conditions, as we design it to deal with complex conditions. 
According to Fig.~\ref{fig-syn}, these state-of-the-art methods cannot remove all rain steaks and may blur the image, while our method can remove the majority of rain steaks as well as maintain details of background scene. 
\begin{figure}[!ht]
	\begin{minipage}{0.24\textwidth}
		\includegraphics[width=\textwidth]{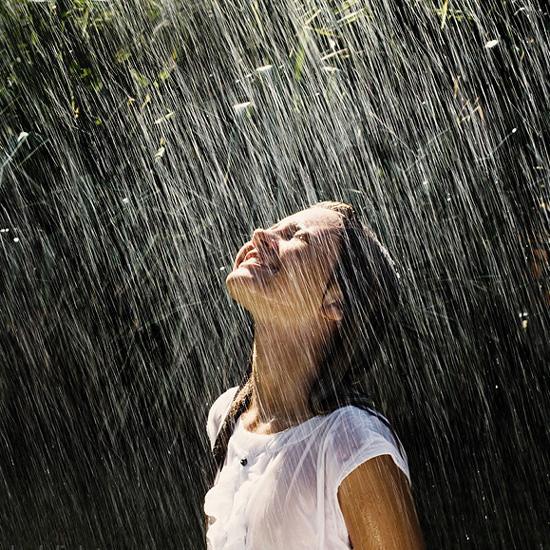}
		\caption*{$\mathbf{O}$}
	\end{minipage}
	\begin{minipage}{0.24\textwidth}
		\includegraphics[width=\textwidth]{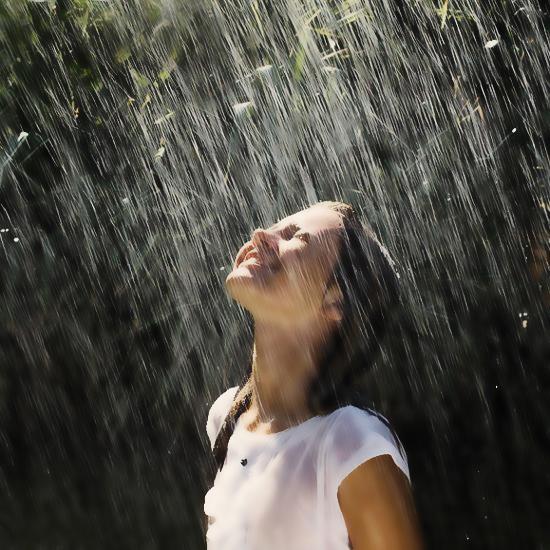}
		\caption*{ID~\cite{kang2012automatic}}
	\end{minipage}
	\begin{minipage}{0.24\textwidth}
		\includegraphics[width=\textwidth]{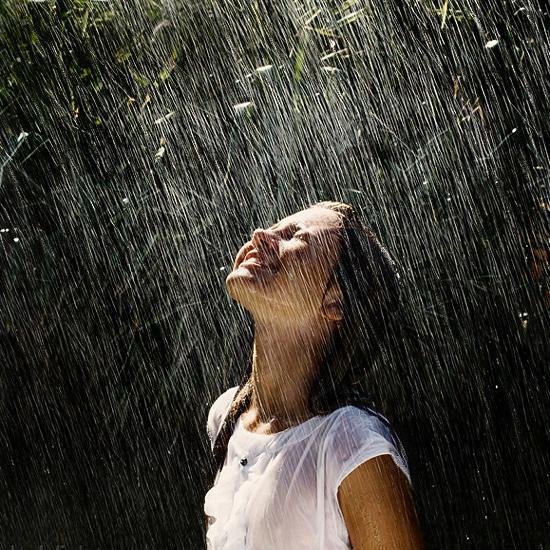}
		\caption*{DSC~\cite{luo2015removing}}
	\end{minipage}
	\begin{minipage}{0.24\textwidth}
		\includegraphics[width=\textwidth]{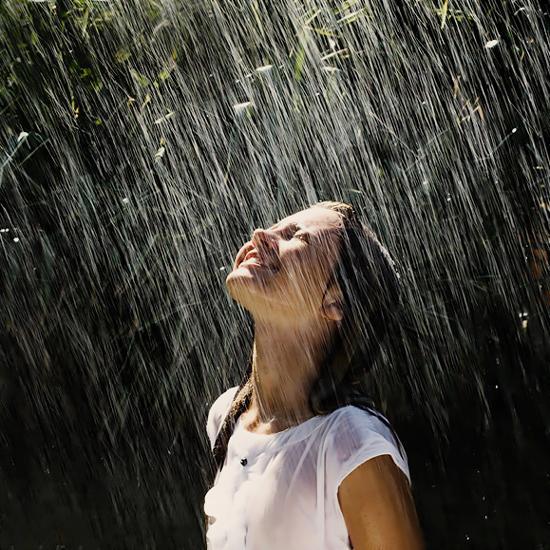}
		\caption*{LP~\cite{li2016rain}}
	\end{minipage}
	\begin{minipage}{0.24\textwidth}
		\includegraphics[width=\textwidth]{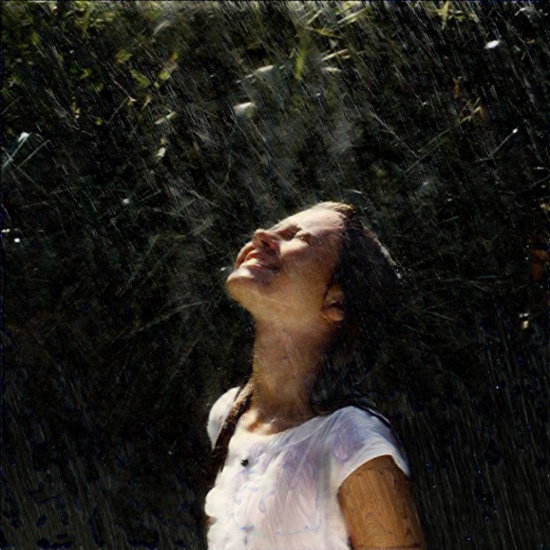}
		\caption*{DetailsNet~\cite{fu2017removing}}
	\end{minipage}
	\begin{minipage}{0.24\textwidth}
		\includegraphics[width=\textwidth]{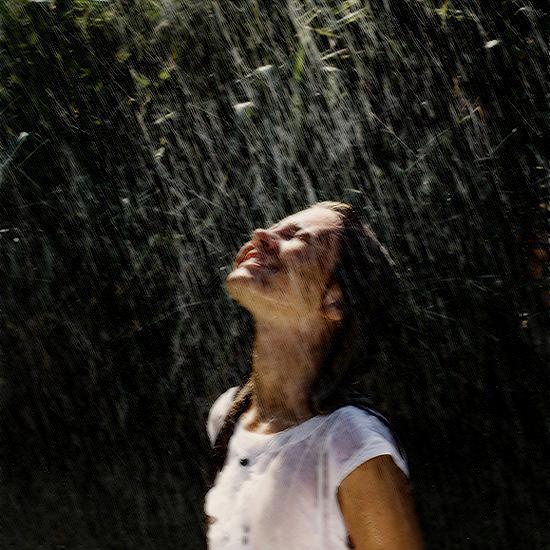}
		\caption*{JORDER-R~\cite{yang2017deep}}
	\end{minipage}
	\begin{minipage}{0.24\textwidth}
		\includegraphics[width=\textwidth]{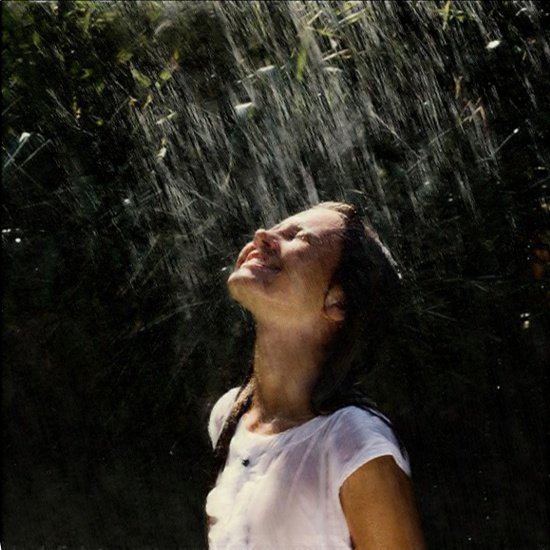}
		\caption*{SCAN}
	\end{minipage}
	\begin{minipage}{0.24\textwidth}
		\includegraphics[width=\textwidth]{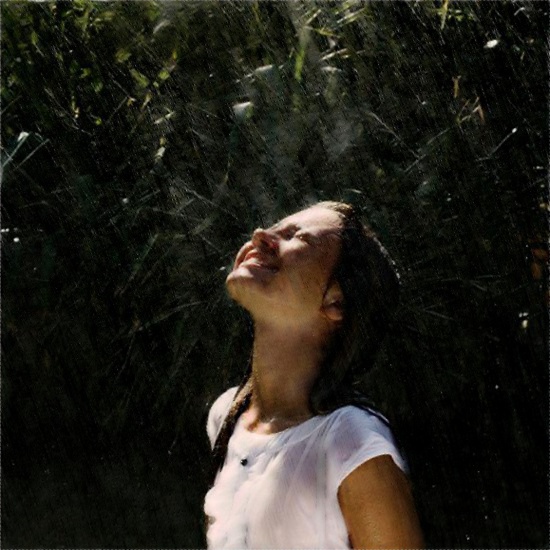}
		\caption*{RESCAN}
	\end{minipage}
	\begin{minipage}{0.24\textwidth}
		\includegraphics[width=\textwidth]{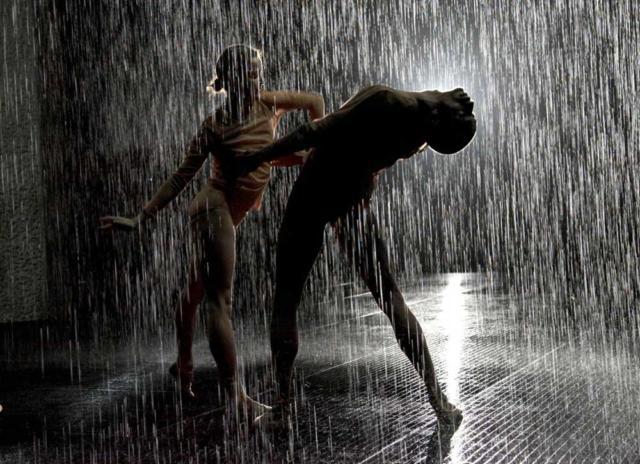}
		\caption*{$\mathbf{O}$}
	\end{minipage}
	\begin{minipage}{0.24\textwidth}
		\includegraphics[width=\textwidth]{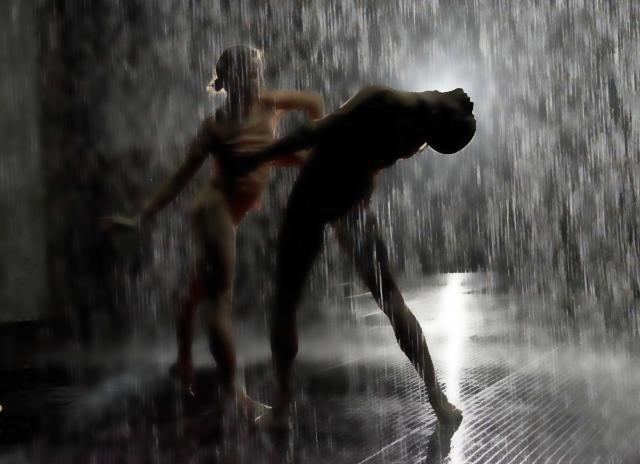}
		\caption*{ID~\cite{kang2012automatic}}
	\end{minipage}
	\begin{minipage}{0.24\textwidth}
		\includegraphics[width=\textwidth]{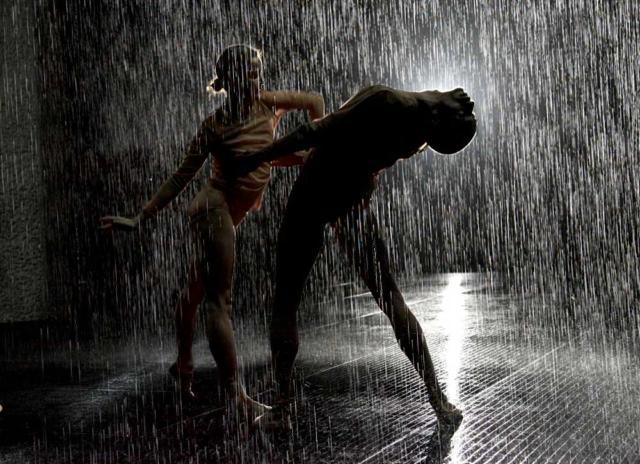}
		\caption*{DSC~\cite{luo2015removing}}
	\end{minipage}
	\begin{minipage}{0.24\textwidth}
		\includegraphics[width=\textwidth]{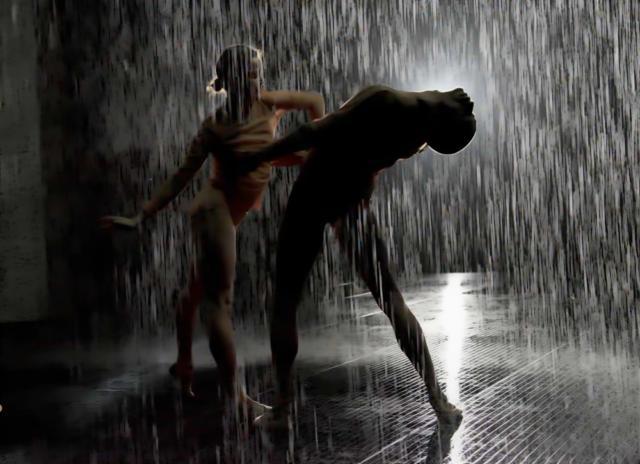}
		\caption*{LP~\cite{li2016rain}}
	\end{minipage}
	
	\begin{minipage}{0.24\textwidth}
		\includegraphics[width=\textwidth]{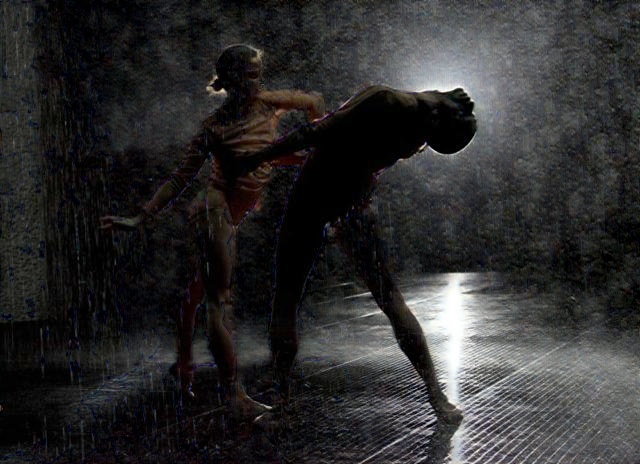}
		\caption*{DetailsNet~\cite{fu2017removing}}
	\end{minipage}
	\begin{minipage}{0.24\textwidth}
		\includegraphics[width=\textwidth]{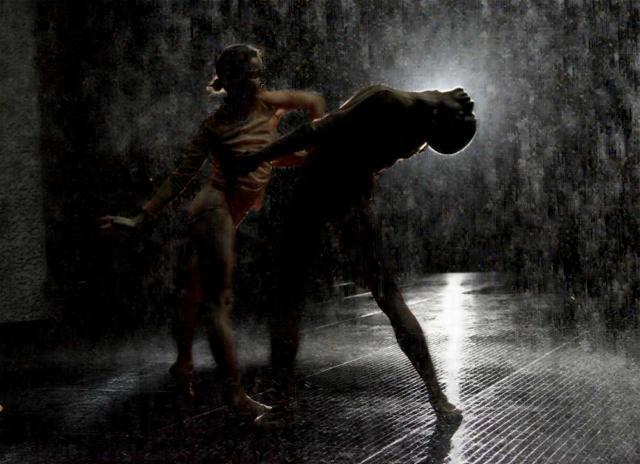}
		\caption*{JORDER-R~\cite{yang2017deep}}
	\end{minipage}
	\begin{minipage}{0.24\textwidth}
		\includegraphics[width=\textwidth]{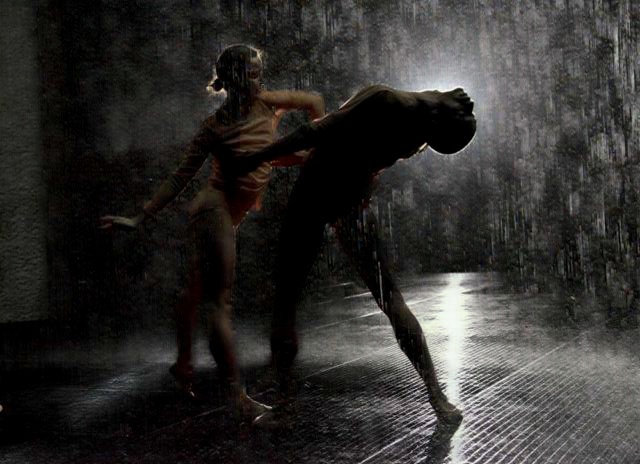}
		\caption*{SCAN}
	\end{minipage}
	\begin{minipage}{0.24\textwidth}
		\includegraphics[width=\textwidth]{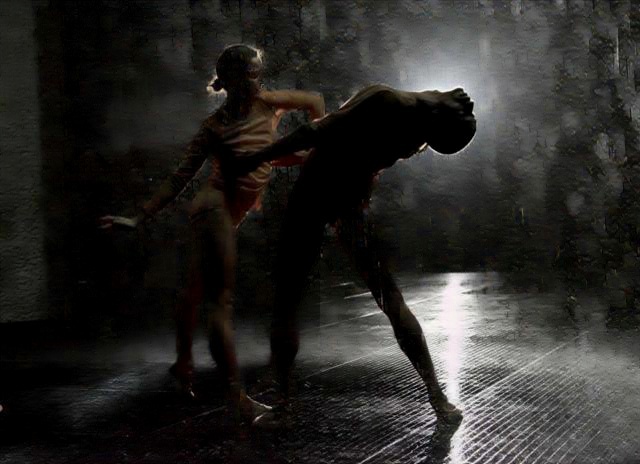}
		\caption*{RESCAN}
	\end{minipage}
	\caption{Results of various methods on real-world images.\! \textbf{Best viewed at screen!}}
	\label{fig-real}
\end{figure}
\subsection{Results on Real-world Dataset}
To test the practicability of deraining methods, we also evaluate the performance on real-world rainy test images. 
The predictions for all related methods on four real-world sample rain images are shown in Fig.~\ref{fig-real}. 
As observed, LP~\cite{li2016rain} cannot remove rain steaks efficiently, and DetailsNet~\cite{fu2017removing} tends to add artifacts on derained outputs. 
We can see that the proposed method can remove most of rain streaks and maintain much texture of background images.
To further validate the performance on real-world dataset, we also make a user study.
For details, please refer to the supplement material.

\subsection{Analysis on SCAN}
To show that SCAN is the best choice as a base model for deraining task, we conduct experiments on two datasets to compare the performance of SCAN and its related network architectures, including Plain~(dilation=$1$ for all convolutions), ResNet used in DetailsNet~\cite{fu2017removing} and Encoder-Decoder used in ID-CGAN~\cite{zhang2017image}. 
For all networks, we set the same depth $d=7$ and width~($24$ channels),
so we can keep their numbers of parameters and computations at the same order of magnitude. 
More specifically, we keep layers $L_0$, $L_{d-5}$ and $L_{d-6}$ of them with the same structure, so they only differ in layers $L_1$ to $L_4$. The results are shown in Table~\ref{table-scan}. 
SCAN and CAN achieve the best performance in both datasets, which can be attributed to the fact that receptive fields of these two methods exponentially increase with the growing of depth, while receptive fields of other methods only have linear relationship with depth.
\par

\begin{table}[t]
	\begin{center}
		\caption{Quantitative comparison between SCAN and base model candidates.}
		\label{table-scan}
		\begin{tabular}{c|c|ccc|cc|cc}
			\hline
			Dataset                & Meassure & Plain  & ResNet & EncDec & CAN+BN & CAN    & SCAN+BN & SCAN        \\
			\hline
			\multirow{2}*{Rain800} & PSNR     & 22.10  & 22.10  & 22.14  & 22.27  & 22.45  & \underline{23.11}   & \textbf{23.45}  \\
			~                      & SSIM     & 0.7816 & 0.7856 & 0.7809 & 0.7871 & \underline{0.7960} & 0.7657  & \textbf{0.8112} \\\hline
			\multirow{2}*{Rain100H}& PSNR     & 21.46  & 21.51  & 21.28  & 22.63  & 22.93  & \underline{23.09}   & \textbf{23.56}  \\
			~                      & SSIM     & 0.6921 & 0.6940 & 0.6886 & 0.7256 & 0.7333 & \underline{0.7389}  & \textbf{0.7456} \\
			\hline
		\end{tabular}
	\end{center}
	\vspace{-2mm}
\end{table}

Moreover, the comparison between results of SCAN and CAN indicates that the SE block contributes a lot to the base model, as it can explicitly learn alpha-value for every independent rain streak layer. We also examine the SCAN model with BN. 
The results clearly verify that removing BN is a good choice in the deraining task, as rain steak layers are independent on each other.

\subsection{Analysis on RESCAN}

As we list three recurrent units and two recurrent frameworks in Section~\ref{sec-rescan}, we conduct experiments to compare these different settings, consisting of $\{$ConvRNN, ConvLSTM, ConvGRU$\}$ $\times$ $\{$Additive Prediction~(Add), Full Prediction~(Full)$\}$. To compare with the recurrent framework in \cite{yang2017deep,li2017single}, we also implement the settings in Eq.~(\ref{eq-iter})~(Iter), in which previous states are not reserved.
In Table~\ref{table-rescan},  we report the results. 

\begin{table}[t]
	\begin{center}
		\caption{Quantitative comparison between different settings of RESCAN. \textit{Iter} represents the framework that leaves out states of previous stages as~\cite{yang2017deep,li2017single}. \textit{Add} and \textit{Full} represent the addtive prediction framework and  the full  prediction framework, respectively.}
		\label{table-rescan}
		\begin{tabular}{ccccc}
			\hline\noalign{\smallskip}
			Dataset  &\multicolumn{2}{c}{Rain800}&\multicolumn{2}{c}{Rain100H} \\
			\noalign{\smallskip}
			\hline
			\noalign{\smallskip}
			Meassure     & PSNR  & SSIM   & PSNR  & SSIM   \\ 
			\noalign{\smallskip}
			\hline
			\noalign{\smallskip}
			SCAN		 & 23.45 & 0.8112 & 23.56 & 0.7456 \\
			Iter+Add     & 23.36 & 0.8169 & 22.81 & 0.7630 \\
			ConvRNN+Add  & 24.09 & 0.8410 & 23.34 & 0.7765 \\
			ConvRNN+Full & 23.52 & 0.8269 & 23.44 & 0.7643 \\
			ConvGRU+Add  & 23.31 & \textbf{0.8444} & 24.00 & 0.7993 \\
			ConvGRU+Full & \underline{24.18} & \underline{0.8394} & \textbf{26.45} & \textbf{0.8458} \\
			ConvLSTM+Add & 22.93 & 0.8385 & 25.13 & 0.8211 \\
			ConvLSTM+Full& \textbf{24.37} & 0.8384 & \underline{25.64} & \underline{0.8334} \\
			\hline
		\end{tabular}
	\end{center}
\end{table}

It is obvious that Iter cannot compete with all RNN structures, and is not better than SCAN, as it leaves out information of previous stages. Moreover, ConvGRU and ConvLSTM outperform ConvRNN, as they maintain more parameters and require more computation. However, it is difficult to pick the best unit between ConvLSTM and ConvGRU as they perform similarly in experiments.
For recurrent frameworks, results indicate that Full Prediction is better.



\section{Conclusions}
We propose the recurrent squeeze-and-excitation based context aggregation network for single image deraining in this paper.
We divide the rain removal into multiple stages.
In each stage, the context aggregation network is adopted to remove rain streaks.
We also modify CAN to better match the rain removal task, including the exponentially increasing dilation and removal of the BN layer.
To better characterize intensities of different rain streak layers, we adopt the squeeze-and-excitation block to assign different alpha-values according to their properties.
Moreover, RNN is incorporated to better utilize the useful information for rain removal in previous stages and guide the learning in later stages.
We also test the performance of different network architectures and recurrent units.
We conduct extensive experiments on both synthetic and real-world datasets, which shows that our proposed method outperforms the
state-of-the-art approaches under all evaluation metrics.
\par
\vspace{3mm}
\noindent
\textbf{Acknowledgment.} \ 
Zhouchen Lin is supported by National Basic Research Program of China~(973 Program)~(Grant no. 2015CB352502), National Natural Science Foundation~(NSF) of China~(Grant nos. 61625301 and 61731018), Qualcomm, and Microsoft Research Asia.
Hong Liu is supported by National Natural Science Foundation of China (Grant nos. U1613209 and 61673030).
Hongbin Zha is supported by Beijing Municipal Natural Science Foundation~(Grant no. 4152006).

\clearpage

\bibliographystyle{splncs}
\bibliography{egbib}
\end{document}